\documentclass[runningheads]{llncs}
\usepackage{graphicx}
\usepackage{bm}
\usepackage{mathtools}
\usepackage{amsmath}
\usepackage{amssymb}
\usepackage{amsfonts}
\usepackage{algorithmic}
\usepackage{multirow}
\usepackage{array}

\usepackage[linesnumbered, ruled, vlined]{algorithm2e}
\usepackage{etoolbox}
\makeatletter
\patchcmd\algocf@Vline{\vrule}{\vrule \kern-0.4pt}{}{}
\patchcmd\algocf@Vsline{\vrule}{\vrule \kern-0.4pt}{}{}
\patchcmd\@makecaption{\scshape}{}{}{}
\patchcmd\@makecaption{\\}{:~}{}{}
\patchcmd\@makecaption{\raggedleft}{}{}{}
\makeatother
\usepackage{changepage}
\usepackage{appendix}
\usepackage{xcolor, colortbl}
\definecolor{Gray}{gray}{0.85}
\usepackage{seqsplit}
\usepackage{wrapfig}
\usepackage[margin=35truemm]{geometry}

\usepackage{hyperref}
\hypersetup{
    colorlinks,
    linkcolor={black},
    citecolor={blue!80!black},
    urlcolor={blue!80!black}
}

\begin{document}
\title{RouteExplainer:\\ An Explanation Framework for Vehicle Routing Problem}
%
%
\author{Daisuke Kikuta* \and Hiroki Ikeuchi \and Kengo Tajiri \and Yuusuke Nakano}
\authorrunning{D. Kikuta et al.}
%
%
\institute{NTT Corporation \\
*\email{daisuke.kikuta@ntt.com}}

\maketitle              
\begin{abstract}
The Vehicle Routing Problem (VRP) is a widely studied combinatorial optimization problem and has been applied to various practical problems. 
While the explainability for VRP is significant for improving the reliability and interactivity in practical VRP applications, it remains unexplored.
In this paper, we propose RouteExplainer, a post-hoc explanation framework that explains the influence of each edge in a generated route. Our framework realizes this by rethinking a route as the sequence of actions and extending counterfactual explanations based on the action influence model to VRP. To enhance the explanation, we additionally propose an edge classifier that infers the intentions of each edge, a loss function to train the edge classifier, and explanation-text generation by Large Language Models (LLMs).
We quantitatively evaluate our edge classifier on four different VRPs. 
The results demonstrate its rapid computation while maintaining reasonable accuracy, thereby highlighting its potential for deployment in practical applications.
Moreover, on the subject of a tourist route, we qualitatively evaluate explanations generated by our framework. This evaluation not only validates our framework but also shows the synergy between explanation frameworks and LLMs.
See \url{https://ntt-dkiku.github.io/xai-vrp} for our code, datasets, models, and demo.
\keywords{Vehicle Routing Problem  \and Explainability \and Structural Causal Model \and Counterfactual Explanation \and Large Language Models.}
\end{abstract}
\section{Introduction}
The Vehicle Routing Problem (VRP) is a combinatorial optimization problem that aims to find the optimal routes for a fleet of vehicles to serve customers.
Since Dantzig and Ramser \cite{vrp_initial} introduced its concept, various VRPs that model real-world problems have been proposed, imposing constraints such as time windows \cite{tsptw_app1}, vehicle capacity \cite{vrp_initial}, and minimum prize \cite{pctsp_app1}. Concurrently, various solvers have been proposed, ranging from exact solvers \cite{generic_bcp} to heuristics \cite{lkh}, Neural Network (NN) solvers \cite{bello,hopfield,ptrnet}, and combinations of them \cite{dpdp,nnpopmusic,neurolkh}.

While existing works have successfully developed various VRPs and their solvers,
explainability for a generated route still remains unexplored. In this paper, we argue that explainability is essential for practical applications such as responsible or interactive route generation.
Furthermore, we argue that explaining how each edge influences the subsequent route is one of the most effective ways to explainability for VRP.
For example, in a route for emergency power supply, 
when asked why the vehicle went to the location instead of other locations at a certain time, a responsible person can justify the decision with the subsequent influence of the movement (i.e., edge).
In interactive tourist route generation, the influence of an edge provides hints to tourists who try to modify, based on their preferences, an edge in an automatically generated route.

By rethinking that a route is created by a chain of cause-and-effects (i.e., actions/movements), we can evaluate the subsequent influence of each edge through causal analysis.
The Structural Causal Model (SCM) \cite{SCM} is one of the most popular models to analyze causality. In SCM, causal dependencies among variables are represented by a Directed Acyclic Graph (DAG).
Recently, to adapt causal analysis to reinforcement learning models, Madumal et al. \cite{causal_rl} introduced the Action Influence Model (AIM), where variables and causal edges are replaced with environment states and actions, respectively. They furthermore proposed a counterfactual explanation based on AIM, which answers \textit{why} and \textit{why-not} questions: why was the action selected instead of other actions at a certain step?

Inspired by this, we propose RouteExplainer, a post-hoc (solver-agnostic) explanation framework that explains the influence of each edge in a generated route.
We modify AIM for VRP (we name it Edge Influence Model (EIM)) considering a route as the sequence of actions (i.e., movements/edges). Based on EIM, our framework generates counterfactual explanations for VRP, which answer why the edge was selected instead of other edges at a specific step.
To enhance the explanation, we additionally incorporate the intentions of each edge as a metric in the counterfactual explanation, and Large Language Models (LLMs) in explanation-text generation.
We here propose a many-to-many sequential edge classifier to infer the intentions of each edge, a modified class-balanced loss function, and in-context learning for LLMs to take in our framework.

In experiments, we evaluate our edge classifier on four different VRP datasets.
Our edge classifier outperforms baselines by a large margin in terms of calculation time while maintaining reasonable accuracy (e.g., 85-90\% in macro-F1 score) on most of the datasets. The results demonstrate its capability for handling a huge number of requests in practical applications. 
Lastly, we qualitatively evaluate counterfactual explanations generated by our framework on a practical tourist route, demonstrating the validity of our framework and the effectiveness of the additional components: intentions of each edge and LLM-powered text generation.

The main contributions of this paper are organized as follows:
\begin{itemize}
    \item[1)] We are the first to argue for the importance of explainability in VRP and propose a novel explanation framework for VRP, including its pipeline and EIM. 
    \item[2)] We propose a many-to-many sequential edge classifier that infers the intention of each edge in a route, which enhances the explanation. 
    \item[3)] To train the edge classifier, we propose a modified class-balanced loss function for step-wise imbalanced classes emerging in our datasets.
    \item[4)] We leverage LLMs to generate the explanation text in our framework, showing the promise of combining explanation frameworks and LLMs.
\end{itemize}

\section{Related Work}
\textbf{Vehicle Routing Problem}\ \ 
The applications of VRP range widely from truck dispatching \cite{vrp_initial} and school bus routing \cite{tsptw_app1} to tourist routing \cite{pctsp_app2}. In each application, we need to select an appropriate VRP solver according to its requirements, including problem size, time limit, accuracy, etc. Today, a variety of solvers exist, ranging from exact solvers to heuristics, neural network (NN) solvers, and their combinations.
Exact solvers such as Branch-Cut-and-Price \cite{generic_bcp} may provide the optimal solution, but its calculation cost is expensive when the problem size is large. 
(Meta-)heuristics such as genetic algorithms and LKH \cite{lkh} provide a near-optimal solution within a reasonable calculation time.
However, they require sophisticated tuning for each VRP to achieve both reasonable calculation time and the quality of routes.
In contrast to these conventional ones, NN solvers with supervised learning \cite{joshi2019efficient,difusco,ptrnet} or reinforcement learning \cite{bello,kool2018attention,nazari,dimes,bert_vrp} have been recently proposed. NN solvers realize faster computation and automatic design of (data-driven) heuristics without domain knowledge for a specific VRP.
More recently, to take advantage of both NNs and heuristics, combinations of the two have emerged \cite{dpdp,nnpopmusic,neurolkh}.
To address this diversification of VRP applications and their solvers, this paper proposes a post-hoc explanation framework that can be applied to any VRP solvers. \vspace{1.5ex plus .2ex minus .2ex}

\noindent\textbf{Explainability}\ \ 
One of the definitions of explainability is the ability to explain the outputs of a model in a way understandable by humans.
We argue that existing VRP solvers lack the explainability:
Conventional solvers inherently possess algorithmic transparency (i.e., the algorithm inside of solvers is known), yet it is difficult to construct interpretable explanations for their outputs by summarizing the complicated optimization processes;
NN solvers are naturally black-box methods, and their outputs are not explainable as is.
In the context of eXplainable Artificial Intelligence (XAI), various explainability methods have been proposed as a proxy between a model and humans.
In particular, post-hoc explainability methods such as feature importance-based methods \cite{lrp,lime,shap} and causal explanations \cite{SCM,causal_rl} are promising to address the explainability for VRP in a solver-agnostic manner.
The former explains the input-output relationships through the future-importance analysis.
The latter, on the other hand, structuralizes the dependencies between features (variables) and explains their importance based on the causal analysis. Considering that the former is limited to NN solvers and the latter provides intermediate cause-and-effect results, the latter is more suitable for VRP.
Recently, Madumal et al. \cite{causal_rl} have adapted causal explanations to actions in reinforcement learning, where variables and directed edges in a DAG are replaced with states and actions, respectively. They approximate structural equations with a simple regression model to construct CF examples and define a pipeline to generate counterfactual explanations. Our framework extends this idea to VRP, where the main differences from \cite{causal_rl} are following: 1) States and actions are replaced with nodes and edges in a route; 2) CF examples are constructed by a VRP solver, and deterministic structural equations are already known; 3) The intention of an edge is additionally considered to improve visual explanation; 4) An LLM is leveraged to generate explanation texts instead of natural language templates.

\section{Proposed Framework: RouteExplainer}
In this section, we introduce the EIM and RouteExplainer.
We here discuss them in terms of the Traveling Salesman Problem with Time Windows (TSPTW).
Given sets of nodes, their positions, and time windows, TSPTW aims to find the shortest route that visits each node exactly once and returns to the original node, where each node must be visited within its time window.

\begin{figure*}[tb] \centering
    \includegraphics[width=\textwidth]{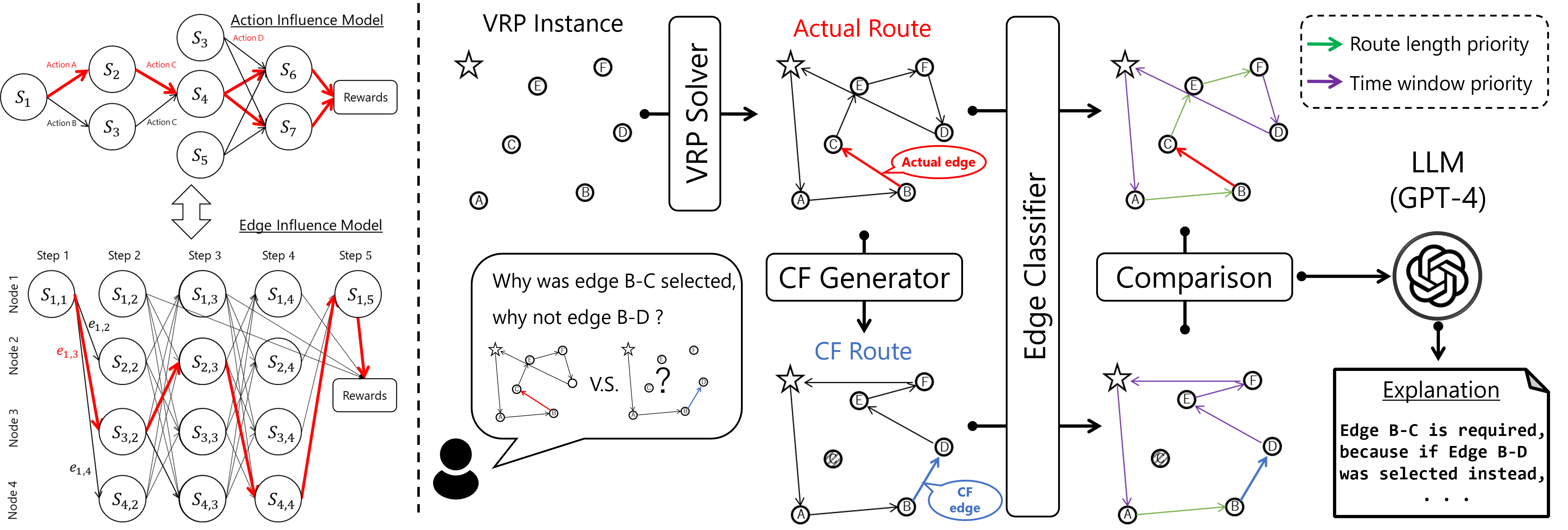}
    \caption{Left: Action Influence Model \cite{causal_rl} v.s. Edge Influence Model (ours). Right: the pipeline of RouteExplainer; it first takes a why and why-not question for VRP and simulates the CF route. The edge classifier then identifies the intentions of each edge in the actual and CF routes, and an LLM (e.g., GPT-4 \cite{openai2023gpt4}) generates a counterfactual explanation by comparing the influences of the actual and CF edges.}
    \label{fig:route_explainer}
\end{figure*}

The l.h.s. of Fig. \ref{fig:route_explainer} shows the comparison between AIM and EIM. In AIM, vertices\footnote{We call "vertices" for nodes in a DAG and "nodes" for nodes in a VRP instance.} are environment states $S_{1,\dots,7}$, and their causal relationships are linked by actions. By contrast, in EIM, vertices are nodes with a \textit{global state} $S_{(i,t)}$ (for TSPTW, intermediate route length/travel time), and their causal relationships are linked by edges of the VRP instance. The bold red path represents a route, which is created by the chain of passed edges. We can intuitively interpret causal relationships in EIM such that the previously visited node affects the global state at the next visited node, e.g., $e_{3,2}$ and $e_{4,2}$ provide different intermediate route lengths at node 2.
The structural equation is $f_{\pi_t\to \pi_{t+1}} = S_{(\pi_{t+1}, t+1)} = S_{(\pi_t,t)} + \textrm{d}(\bm{x}_{v_{\pi_t}}, \bm{x}_{v_{\pi_{t+1}}})$, where $\textrm{d}(\cdot, \cdot)$ is a function that computes the distance between two nodes.
Counterfactual explanations are generated based on EIM.

The r.h.s of Fig. \ref{fig:route_explainer} shows the pipeline of RouteExplainer.
Given an automatically generated route (we call it \textit{actual route}), we consider a situation where a user asks why edge B-C was selected at the step and why not edge B-D, based on the user's thought or preference.
Our framework answers this question by comparing the influence of the two edges.
It first takes this question and simulates a counterfactual (CF) route, which is an alternative route where edge B-D is instead selected at the step. Then, the edge classifier infers the intentions of edges in both the actual and CF routes.
Finally, an LLM generates counterfactual explanations for comparing the influence of edges B-C and B-D.

Hereafter, we describe the edge classifier and explanation generation in detail. 
\vspace{1.5ex plus .2ex minus .2ex}

\noindent\textbf{Notation}
Let $\mathcal{G} = (\mathcal{V}, \mathcal{E}, X)$ be an undirected graph of a VRP instance, where nodes $v_i\in\mathcal{V}; i\in\{1,...,N\}$ are destinations, $\bm{x}_{v_i}\in X$ is the node feature, edges $e_{ij}\in\mathcal{E}$ are feasible movements between $v_i$ and $v_j$; $i\neq j$, and $N(=|\mathcal{V}|)$ is the number of nodes. 
In this paper, we consider only complete graphs, i.e., all movements between any two nodes are feasible.
Given the visiting order of nodes in a route $\bm{\pi}=(\pi_1, ..., \pi_T)$, 
the route is represented by the sequence of edges $\bm{e} = (e_1, ..., e_{T-1})=(e_{\pi_1\pi_2}, ..., e_{\pi_{T-1}\pi_T})$, where $\pi_t\in\{1,...,N\}$ is the index of the node visited at step $t$ and $T$ is the sequence length of the route. 
\subsection{Many-to-many Edge Classifier}
\begin{wrapfigure}[14]{r}[0pt]{0.5\textwidth}
    \vspace{-25pt}
    \centering
    \includegraphics[width=0.5\textwidth]{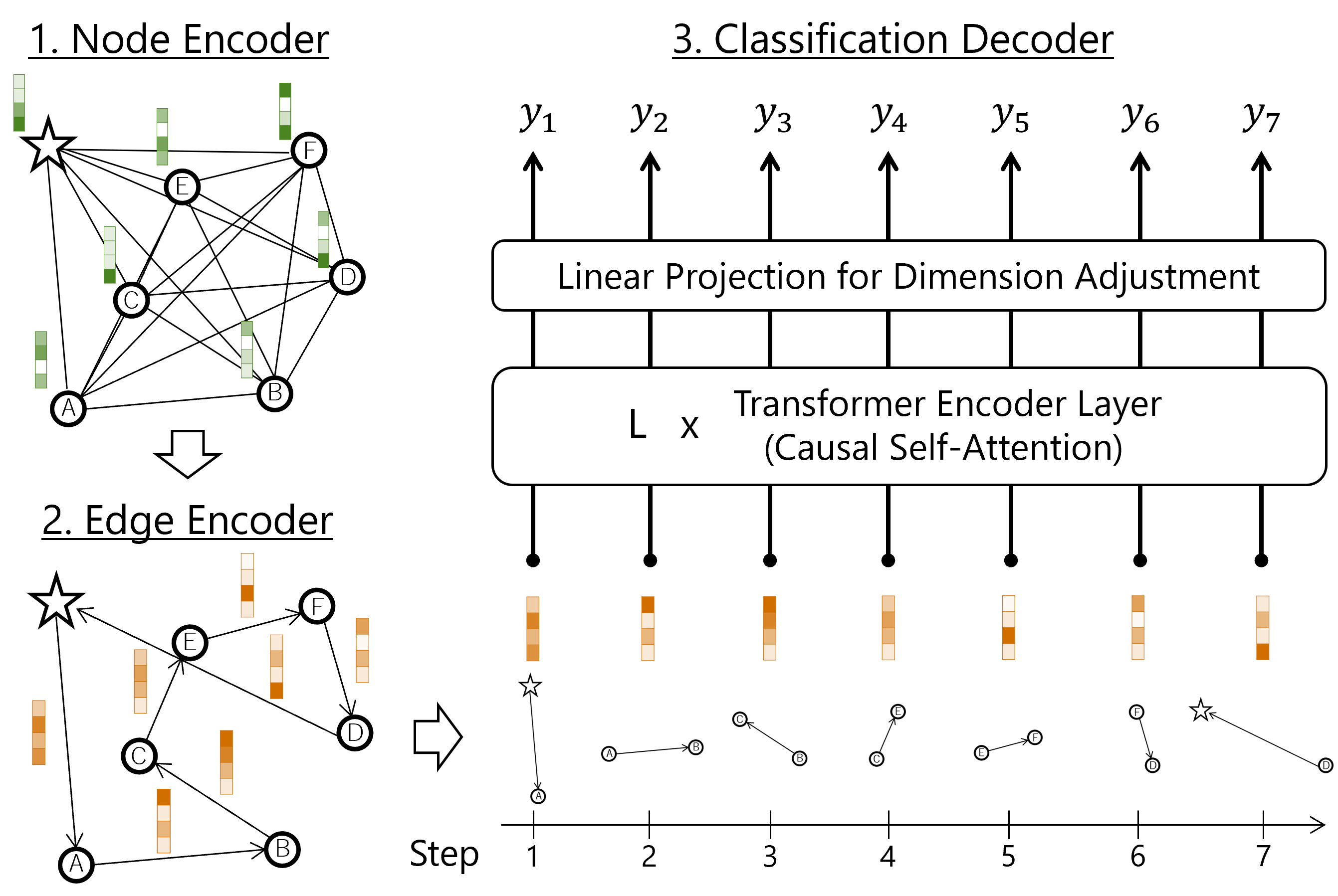}
    \vspace{-10pt}
    \caption{The proposed many-to-many edge classifier.} 
    \label{fig:edge_classifier}
\end{wrapfigure} 
Assuming that each edge in the routes has a different intention of either prioritizing route length or time constraint in a TSPTW route, the edge classifier aims to classify the intentions of each edge.
We formulate this as a many-to-many sequential classification, i.e., the edge classifier takes a route (sequence of edges) and sequentially classifies each edge.
We here propose a Transformer-based edge classifier, which consists of a node encoder, edge encoder, and classification decoder (Fig. \ref{fig:edge_classifier}).
In the following, we describe the details of each component and how to train the classifier in supervised learning.\vspace{1.5ex plus .2ex minus .2ex}

\noindent\textbf{Node Encoder}\ \ 
The node encoder aims to convert input node features into higher-dimensional representations that consider the dependencies with other nodes. 
As in \cite{kool2018attention}, we here employ the Transformer encoder by Vaswani et al. \cite{transformer}, but without the positional encoding since nodes are permutation invariant.
First, the $i$-th node's input features $\bm{x}_{v_i}\in\{\mathbb{R}^{D'}, \mathbb{R}^D\}$ is projected to node embeddings $\bm{h}_{v_i}\in\mathbb{R}^H$
with different linear layers for the depot and other nodes.
\begin{equation}
    \bm{h}^{(0)}_{v_i} = \left\{
    \begin{array}{ll}
        W_{\textrm{depot}}\bm{x}_{v_i} + \bm{b}_{\textrm{depot}} & i=1,\\
        W\bm{x}_{v_i} + \bm{b} & i=2,\ldots,N,
    \end{array}
    \right.
\end{equation} 
where $W_{\textrm{depot}}\in\mathbb{R}^{H\times D'}, W\in\mathbb{R}^{H\times D}$, $\bm{b}_\textrm{depot}, \bm{b}\in\mathbb{R}^H$ are projection matrices and biases for the depot and other nodes, respectively.
See Appendix \ref{appendix:data_generation} for details of input features.
Then we obtain the final node embeddings by stacking $L$ Transformer encoder layers \textsc{Xfmr}, as follows:
\begin{equation}
    \bm{h}_{v_i}^{(l)}=\textsc{Xfmr}_{v_i}^{(l)}(\bm{h}^{(l-1)}_{v_1},\cdots,\bm{h}^{(l-1)}_{v_N}),
\end{equation}
where $l$ indicates $l$-th layer.\vspace{1.5ex plus .2ex minus .2ex}

\noindent\textbf{Edge Encoder}\ \ 
The edge encoder generates edge embeddings for the edges in the input route.
For the embedding of the edge at step $t$, it simply concatenates the final node embeddings of both ends of the edge and the global state at that step.
\begin{equation}
    \label{eq:edge_encoder}
    \bm{h}^{(0)}_{e_t} = W_\textrm{edge}\left[\bm{h}^{(L)}_{v_{\pi_t}} || \bm{h}^{(L)}_{v_{\pi_{t+1}}}||\bm{s}\right],
\end{equation}
where $\bm{s}\in\mathbb{R}^{D_\text{st}}$ is the environment state value, $||$ indicates concatenation w.r.t. feature dimensions, and $W_\textrm{edge}\in\mathbb{R}^{H\times (2H+D_\text{st})}$ is the projection matrix. \vspace{1.5ex plus .2ex minus .2ex}

\noindent\textbf{Classification Decoder}\ \ 
The decoder takes the sequence of the edge embeddings and outputs probabilities of each edge being classified into classes.
Similar to the node encoder, we here employ the Transformer encoder, but with causal masking, i.e., it considers only the first $t$ edges when computing the embedding of the edge at step $t$. 
This causality ensures the consistency of the predicted labels of common edges in the actual and CF routes. 
In addition, we do not use any positional encodings as positional information is already included in the environment state value $\bm{s}$.
We obtain the new edge embeddings by stacking $L'$ Transformer encoder layers as follows:
\begin{equation}
    \bm{h}^{(l)}_{e_t} = \textsc{Xfmr}^{(l)}_{e_t}(\bm{h}^{(l-1)}_{e_1},\ldots,\bm{h}^{(l-1)}_{e_{t}}).
\end{equation}
Finally, we obtain the probabilities of the edge at step $t$ being classified into the classes of intentions with a linear projection and the softmax function. 
\begin{equation}
    \bm{p}_{e_t} = \textrm{Softmax}\left(W_{\textrm{dec}}\bm{h}^{(L')}_{e_t} + \bm{b}_{\textrm{dec}}\right),
\end{equation}
where $W_{\textrm{dec}}\in\mathbb{R}^{C\times{H}}, \bm{b}_{\textrm{dec}}\in\mathbb{R}^C$ are the projection and bias that adjust the output dimension to the number of classes $C$. In inference, the predicted class is determined by the argmax function, i.e., $\hat{y}_{e_t}=\textrm{argmax}_c(\bm{p}_{e_t})$.\vspace{1.5ex plus .2ex minus .2ex}

\noindent\textbf{Training}\ \ 
For supervised learning, we need labels of edges in routes.
In this paper, we employ a rule-based edge annotation for simplicity and to remove human biases (see Appendix \ref{appendix:data_generation} for the details of annotation).. 
Notably, the advantage of machine learning models is that they can also accommodate manually annotated data.
We train the edge classifier with the generated labels. 
A challenge in our problem setting is that the class ratio changes over each step.
In TSPTW, we observe a transitional tendency such that time window priority is the majority class in the early steps, whereas route length priority becomes the majority as the steps progress.
Usually, a weighted loss function and class-balanced loss function \cite{CBCE} are used for class-imbalanced data.
However, they only consider the class ratio in the entire training batch, which fails to capture step-wise class imbalances as seen in TSPTW datasets.
Therefore, we propose a cross-entropy loss that considers step-wise class imbalance. 
We adjust the class-balanced loss to many-to-many sequential classification by calculating the weights separately at each step, as follows,
\begin{equation}
    \label{eq:seq_cbce}
    J = - \sum_{T\in\mathcal{T}}\sum_{b=1}^{B_T}{\sum_{t=1}^T\sum_{c=0}^{C-1}\frac{1-\beta}{1-\beta^{\sum_b\textrm{I}\left(y_{e^b_t}=c\right)}}{y_{e^b_t}\log{\left((\bm{p}_{e^b_t})_c\right)}}},
\end{equation}
where $y_{e^b_t}$ is the true label of the edge at $t$ in the route of the instance $b$, $\beta$ is a hyperparameter that corresponds to the ratio of independent samples, $\textrm{I}(\cdot)$ is the Boolean indicator function, $\mathcal{T}$ is the set of sequence lengths included in the training batch, $B_T$ is the number of samples whose sequence length is $T$, and $(\bm{p}_{e^b_t})_c$ indicates the $c$-th element of $\bm{p}_{e^b_t}$.

\subsection{Counterfactual Explanation for VRP}
In this section, we describe the explanation generation by defining why and why-not questions in VRP, CF routes, and counterfactual explanations.\vspace{1.5ex plus .2ex minus .2ex}

\noindent\textbf{Why and Why-not Questions in VRP}\ \ 
In the context of VRP, a why and why-not question asks \textit{why edge B-C was selected at the step and why not edge B-D}.
Formally, we define the why and why-not question in VRP as a set:
\begin{equation}
    \label{eq:wwn}
    \mathcal{Q}=(\bm{e}^\text{fact},t_\text{ex}, e^\text{fact}_{t_\text{ex}}, e^\textrm{cf}_{t_\text{ex}}),
\end{equation}
where $\bm{e}^\text{fact}$ is the actual route, $t_\text{ex}\in\{1,\ldots,T-1\}$ is the explained step, $e^\text{fact}_{t_\textrm{ex}}$ is the actual edge, and $e^\textrm{cf}_{t_\textrm{ex}}(\neq e^\text{fact}_{t_\textrm{ex}})$ is the CF edge, which is an edge that was not selected at $t_\text{ex}$ in the actual route but could have been (i.e., \textit{edge B-D} in the above example).\vspace{1.5ex plus .2ex minus .2ex}

\noindent\textbf{Counterfactual Routes}\ \ 
Given a why and why-not question $\mathcal{Q}$, our framework generates its CF route.
The CF route $\bm{e}^{\mathcal{Q}}$ is the optimal route that includes both the sequence of edges before the explained step in the actual route and the CF edge:
\begin{align}
    \bm{e}^{\mathcal{Q}} &= \textsc{Solver}\left(\textsc{Vrp}, \mathcal{G}, \bm{e}_\text{fixed}\right), \\
    \bm{e}_\text{fixed} &= (e^\text{fact}_1,\ldots,e^\text{fact}_{(t_{\textrm{ex}}-1)},e^\textrm{cf}_{{t_\textrm{ex}}}),
\end{align}
where $\bm{e}^\mathcal{Q}[\colon t_{\textrm{ex}}]=\bm{e}^\textrm{fact}[\colon t_{\textrm{ex}}]$, $\bm{e}[\colon{t'}]\coloneqq(e_1,\ldots,e_{t'-1})$,
$e^\mathcal{Q}_{t_\text{ex}} = e^\textrm{cf}_{t_\text{ex}}$, \textsc{Vrp} is the VRP where the actual route was solved, and \textsc{Solver} is the VRP solver.
Intuitively, the CF route simulates the best-effort route for the remaining subsequent steps when a CF edge is selected instead of the actual edge at the explained step.\vspace{1.5ex plus .2ex minus .2ex}

\noindent\textbf{Explanation Generation}\ \ 
To evaluate the influence of an edge, we leverage the global state of visited nodes in EIM, the intentions of each edge, and other metrics such as feasibility ratio.
Given a route $\bm{\pi}=(\pi_1,\ldots,\pi_{T})$ and the intentions of edges in the route $\bm{\hat{y}}=(\hat{y}_1,\ldots,\hat{y}_{T-1})$,
we generally define the influence of the edge $e_t$ as a tuple:
\begin{equation}
    \label{eq:general_influence}
    \bm{I}_{e_t}=\left(S_{(\pi_{t+1}, t+1)}, \hat{y}_{t+1},S_{(\pi_{t+2}, t+2)}, \ldots, \hat{y}_{T-1}, S_{(\pi_{T}, T)}\right).
\end{equation}
Furthermore, we shall construct a \textit{minimally complete explanation} \cite{causal_rl} so that one can effectively understand the influence of an edge.
We here use representative values calculated from Eq. (\ref{eq:general_influence}), which include objective values, the ratio of each class, the feasibility ratio, and so on:
\begin{equation}
    \mathcal{R}_{e_t} = \mathcal{F}_\text{rep}(\bm{I}_{e_t}),
\end{equation}
where $\mathcal{R}_{e_t}$ is the set of representative values and $\mathcal{F}_\text{rep}$ is the set of representative value functions.
For TSPTW, $\mathcal{F}_\text{rep}=(f^\textrm{s}_\text{obj}, f^\textrm{l}_\text{obj}, f_\text{class}, f_\text{feas})$, where the short-term objective value $f^\textrm{s}_\text{obj}(\bm{I}_{e_t})=S_{(\pi_{t+1}, t+1)}$, the long-term objective value $f^\textrm{l}_\text{obj}(\bm{I}_{e_t})=S_{(\pi_{T}, T)}$, the class ratio $f_\text{class}(\bm{I}_{e_t})= \frac{1}{T-t-1}\sum_{t'=t+1}^{T-1}\textrm{I}({\hat{y}_{t'}=0})$, and the feasibility ratio $f_\text{feas}(\bm{I}_{e_t})=\frac{|\{v_{\pi_{t+1}}, \dots, v_{\pi_T}\}|}{|\mathcal{V}\backslash \{v_{\pi_1}, \dots, v_{\pi_t}\}|}$.

The counterfactual explanation is generated by comparing the representative values of actual and CF edges as follows,
\begin{equation}
    \label{eq:explanation}
    \mathcal{X}_\mathcal{Q}=\mathcal{F}_\text{compare}\left(\mathcal{R}_{e^\text{fact}_{t_\text{ex}}}, \mathcal{R}_{e^\mathcal{Q}_{t_\text{ex}}}\right),
\end{equation}
where $\mathcal{X}_\mathcal{Q}$ is the counterfactual explanation for $\mathcal{Q}$, $\mathcal{R}_{e^\text{fact}_{t_\text{ex}}}, \mathcal{R}_{e^\mathcal{Q}_{t_\text{ex}}}$ are the sets of representative values of actual and CF edges, respectively, and $\mathcal{F}_\text{compare}$ is the function that compares the two sets of representative values (we here use element-wise difference operation). 
Finally, we generate an explanation text with the compared representative values using GPT-4 \cite{openai2023gpt4}, an LLM. NLP templates require complicated conditional branches to generate user-friendly explanations, whereas LLMs realize this only with simple natural language instructions. We incorporate our framework into GPT-4 by writing the description of our framework and an example of explanation text in the input context (i.e., In-context learning). See Appendix \ref{appendix:llm} for details of GPT-4 configurations, including the system architecture and system prompt. In practice, we also leverage visualization information, which is demonstrated in the experiments.

\section{Experiments\protect\footnote{Our code is publicly available at \url{https://github.com/ntt-dkiku/route-explainer/}.}}
\begin{wrapfigure}[13]{r}[0pt]{0.5\textwidth}
    \centering
    \vspace{-25pt}
    \includegraphics[width=0.5\textwidth, height=0.25\textwidth]{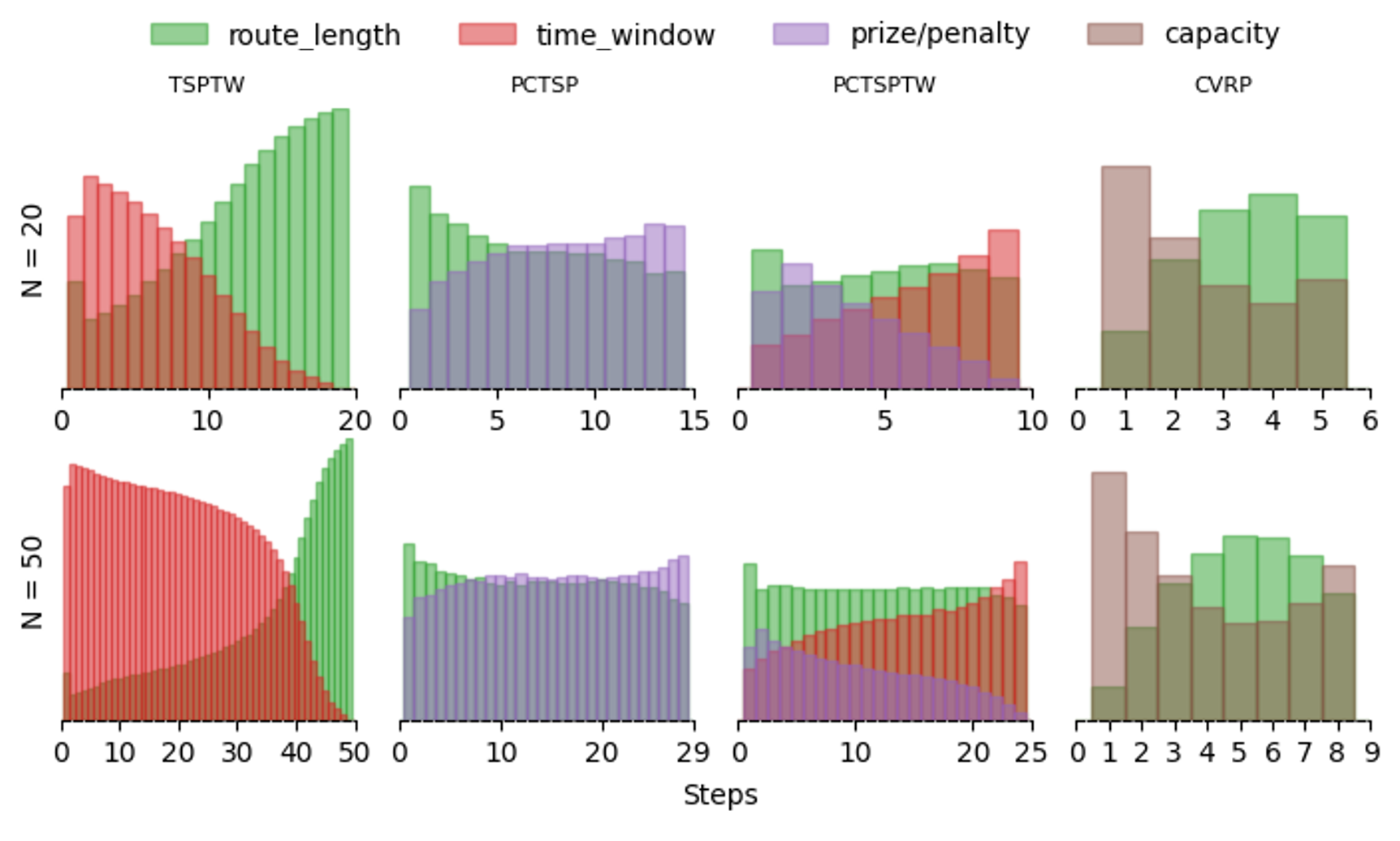}
    \caption{The class ratio of edges w.r.t. steps, on each training split. 
    The class ratios are for samples in which the number of visited nodes is the mode.} 
    \label{fig:statistics}
\end{wrapfigure}
\textbf{Datasets}\ \ 
We evaluate our edge classifier on \textsc{\seqsplit{Actual-Route-Datasets}} and \textsc{\seqsplit{CF-Route-Datasets}}.
Each includes routes and their edges' labels in TSPTW, Prize Collecting TSP (PCTSP), PCTSP with Time Windows (PCTSPTW), and Capacitated VRP (CVRP) with $N=20, 50$.
The labels are annotated by the rule-based annotation in the previous section.
\textsc{\seqsplit{Actual-Route-Datasets}} are split into training, validation, and test splits.
Fig. \ref{fig:statistics} shows the statistics of the training split in each VRP of \textsc{Actual-Route-Datasets}.
See Appendix \ref{appendix:data_generation} for the details of each VRP dataset.\vspace{1.5ex plus .2ex minus .2ex}

\noindent\textbf{Baselines}\ \ 
As baselines against our edge classifier, we use the annotation strategy with different VRP solvers, including LKH \cite{lkh}, Concorde\footnote{\url{https://www.math.uwaterloo.ca/tsp/concorde/}}, and Google OR Tools\footnote{\url{https://github.com/google/or-tools/}} (one of them corresponds to the ground truth).
Note that the baselines are limited to classification labeled by the rule-based annotation, whereas our edge classifier is capable of handling classification labeled by other strategies such as manual annotation. See Appendix \ref{appendix:baslines} for the details of the baselines.

\subsection{Quantitative Evaluation of the Edge Classifier}
\begin{table*}[tb]
    \caption{Macro-F1 (MF1) and calculation time (Time) on \textsc{Actual-Route-Datasets} and \textsc{CF-Route-Datasets} (10K evaluation samples). The ground truth is grayed out.}
    \begin{center}
    \footnotesize
    \begin{tabular}{l@{\extracolsep{1.8pt}} c@{\extracolsep{1.8pt}}cc@{\extracolsep{1.8pt}}c c@{\extracolsep{1.8pt}}cc@{\extracolsep{1.8pt}}c c@{\extracolsep{1.8pt}}cc@{\extracolsep{1.8pt}}c c@{\extracolsep{1.8pt}}cc@{\extracolsep{1.8pt}}c}
        \hline 
        &\multicolumn{16}{c}{\textsc{Actual-Route-Datasets}} \\
        \cline{2-17}
        &\multicolumn{4}{c}{TSPTW} &\multicolumn{4}{c}{PCTSP}  &\multicolumn{4}{c}{PCTSPTW} &\multicolumn{4}{c}{CVRP}\\
        \cline{2-5}\cline{6-9}\cline{10-13}\cline{14-17}
        &\multicolumn{2}{c}{$N=20$} &\multicolumn{2}{c}{$N=50$} &\multicolumn{2}{c}{$N=20$} &\multicolumn{2}{c}{$N=50$} &\multicolumn{2}{c}{$N=20$} &\multicolumn{2}{c}{$N=50$} &\multicolumn{2}{c}{$N=20$} &\multicolumn{2}{c}{$N=50$}\\
        \cline{2-3}\cline{4-5}\cline{6-7}\cline{8-9}\cline{10-11}\cline{12-13}\cline{14-15}\cline{16-17}
        &MF1 &Time &MF1 &Time &MF1 &Time &MF1 &Time &MF1 &Time &MF1 &Time &MF1 &Time &MF1 &Time \\
        \hline
        LKH      &\cellcolor{Gray}100  &\cellcolor{Gray}(2m)  &\cellcolor{Gray}100  &\cellcolor{Gray}(4m) &\cellcolor{Gray}100  &\cellcolor{Gray}(1m)  &\cellcolor{Gray}100  &\cellcolor{Gray}(2m) &- &- &- &- &\cellcolor{Gray}100 &\cellcolor{Gray}(2m) &\cellcolor{Gray}100 &\cellcolor{Gray}(5m)\\
        Concorde &\textbf{100}  &(2m)  &\textbf{99.9} &(8m) &\textbf{99.9} &(2m)  &\textbf{99.6} &(3m) &- &- &- &- &90.5 &(3m) &\textbf{96.3} &(12m)\\
        ORTools  &96.5 &(16s) &94.3 &(4m) &93.5 &(14s) &84.9 &(3m) &\cellcolor{Gray}100 &\cellcolor{Gray}(25s) &\cellcolor{Gray}100 &\cellcolor{Gray}(7m) &\textbf{93.5} &(24s) &89.1 &(6m)\\
        \cline{1-17}
        EC$^{\text{--enc}}_{\text{scbce}}$ &\textbf{93.1}  &(1s) &\textbf{92.0} &(1s) &78.4 &(1s) &75.5 &(1s) &66.8 &(1s) &60.8 &(1s) &83.5 &(2s) &82.8 &(2s) \\
        EC$^{\text{--dec}}_{\text{cbce}}$  &90.9  &(1s) &90.8 &(1s) &\textbf{84.4} &(1s) &\textbf{79.2} &(1s) &\textbf{76.1} &(1s) &\textbf{65.6} &(1s) &\textbf{88.6} &(2s) &\textbf{87.1} &(2s)\\
        \hline
        EC$_{\text{ce}}$                   &\textbf{94.5}  &(1s) &\textbf{93.1} &(1s) &\textbf{90.2} &(1s) &\textbf{84.8} &(1s) &78.4 &(1s) &\textbf{69.2} &(1s) &\textbf{91.1} &(2s) &\textbf{88.9} &(3s)\\
        EC$_{\text{cbce}}$                 &\textbf{94.5}  &(1s) &\textbf{93.1} &(1s) &89.9 &(1s) &84.4 &(1s) &\textbf{78.7} &(1s) &69.0 &(1s) &\textbf{91.1} &(2s) &\textbf{88.9} &(3s)\\
        EC$_{\text{scbce}}$                &94.2  &(1s) &92.5 &(1s) &89.9 &(1s) &84.7 &(1s) &76.6 &(1s) &67.4 &(1s) &90.5 &(2s) &88.2 &(3s) \\
        \hline \hline
        &\multicolumn{16}{c}{\textsc{CF-Route-Datasets}} \\
        \cline{2-17}
        &\multicolumn{4}{c}{TSPTW} &\multicolumn{4}{c}{PCTSP}  &\multicolumn{4}{c}{PCTSPTW} &\multicolumn{4}{c}{CVRP}\\
        \cline{2-5}\cline{6-9}\cline{10-13}\cline{14-17}
        &\multicolumn{2}{c}{$N=20$} &\multicolumn{2}{c}{$N=50$} &\multicolumn{2}{c}{$N=20$} &\multicolumn{2}{c}{$N=50$} &\multicolumn{2}{c}{$N=20$} &\multicolumn{2}{c}{$N=50$} &\multicolumn{2}{c}{$N=20$} &\multicolumn{2}{c}{$N=50$}\\
        \cline{2-3}\cline{4-5}\cline{6-7}\cline{8-9}\cline{10-11}\cline{12-13}\cline{14-15}\cline{16-17}
        &MF1 &Time &MF1 &Time &MF1 &Time &MF1 &Time &MF1 &Time &MF1 &Time &MF1 &Time &MF1 &Time \\
        \hline
        LKH      &\cellcolor{Gray}100  &\cellcolor{Gray}(2m)  &\cellcolor{Gray}100  &\cellcolor{Gray}(4m) &\cellcolor{Gray}100  &\cellcolor{Gray}(1m)  &\cellcolor{Gray}100  &\cellcolor{Gray}(3m) &- &- &- &- &\cellcolor{Gray}100 &\cellcolor{Gray}(2m) &\cellcolor{Gray}100 &\cellcolor{Gray}(5m)\\
        Concorde &\textbf{99.7} &(2m)  &\textbf{99.9}  &(7m) &\textbf{100}  &(2m)  &\textbf{99.6} &(6m) &- &- &- &- &90.8 &(3m) &\textbf{96.2} &(10m)\\
        ORTools  &96.8 &(15s) &94.6  &(4m) &\textbf{93.9} &(14s) &85.7 &(4m) &\cellcolor{Gray}100 &\cellcolor{Gray}(24s) &\cellcolor{Gray}100 &\cellcolor{Gray}(6m) &94.0 &(25s) &89.5 &(6m)\\
        \cline{1-17}
        EC$^{\text{--enc}}_{\text{scbce}}$ &\textbf{91.3}  &(1s) &90.2 &(1s) &78.2 &(1s) &76.0 &(1s) &66.1 &(1s) &62.0 &(1s) &80.4 &(2s) &81.5 &(3s)\\
        EC$^{\text{--dec}}_{\text{cbce}}$  &88.8  &(1s) &\textbf{90.6} &(1s) &\textbf{83.9} &(1s) &\textbf{80.0} &(1s) &\textbf{73.9} &(1s) &\textbf{66.7} &(1s) &\textbf{86.4} &(2s) &\textbf{86.3} &(2s)\\
        \hline
        EC$_{\text{ce}}$                   &\textbf{93.2}  &(1s) &\textbf{92.2} &(1s) &\textbf{88.2} &(1s) &84.1 &(1s) &76.3 &(1s) &\textbf{69.6} &(1s) &89.1 &(2s) &88.5 &(3s)\\
        EC$_{\text{cbce}}$                 &93.1  &(1s) &91.9 &(1s) &88.1 &(1s) &\textbf{84.4} &(1s) &\textbf{76.9} &(1s) &\textbf{69.6} &(1s) &\textbf{89.3} &(2s) &\textbf{88.6} &(3s)\\
        EC$_{\text{scbce}}$                &92.7  &(1s) &92.0 &(1s) &88.0 &(1s) &84.2 &(1s) &75.2 &(1s) &67.9 &(1s) &\textbf{89.3} &(2s) &88.0 &(3s)\\
        \hline
    \end{tabular}
    \label{tab:results}
    \end{center}
\end{table*}
Table \ref{tab:results} shows the quantitative results in classification on \textsc{\seqsplit{Actual-Route-Datasets}} and \textsc{\seqsplit{CF-Route-Datasets}}: the macro f1 score (MF1) and total inference time (Time). 
Here, we report the results of baselines and ablations of our model: EC$^{\text{--enc}}_{\text{scbce}}$/EC$^{\text{--dec}}_{\text{cbce}}$ is the edge classifier that replaces the Transformer encoder in the node encoder/decoder with Multi-Layer Perceptron (MLP). We also report the variants of our model: EC$_{\text{ce}}$, EC$_{\text{cbce}}$, and EC$_{\text{scbce}}$. The subscript indicates the loss function: cross-entropy loss (CE), class-balanced CE (CBCE), and step-wise CBCE (SCBCE, Eq(\ref{eq:seq_cbce})).
All the results of our model are from the model with the best epoch on the validation split. See Appendix \ref{appendix:hyperparameter} for the hyperparameters. 
In the following, we discuss the performance comparison and ablation study.\vspace{1.5ex plus .2ex minus .2ex}

\noindent\textbf{Performance Comparison}\ \
We here focus on the variants of our model: EC$_{\text{ce}}$, EC$_{\text{cbce}}$, and EC$_{\text{scbce}}$. EC$^{\text{--enc}}_{\text{scbce}}$ and EC$^{\text{--dec}}_{\text{cbce}}$ are discussed in ablation study section. 
For \textsc{Actual-Route-Datasets}, our models significantly improve the inference time while retaining 85-95\% of MF1 on most VRPs. In PCTSPTW, MF1 remains around 69-78\%, but it is the only three-class classification and not so low compared to random classification (i.e., 33.3\%).
Real-time response is essential for practical applications.
The baselines require more than 3 minutes for 10K samples, whereas our models require less than 3 seconds, thereby demonstrating their potential for rapidly handling a huge number of requests in practical applications.
For \textsc{CF-Route-Datasets}, the results are similar to those in \textsc{Actual-Route-Datasets}.
Comparing MF1 in both datasets, our models reduce MF1 in \textsc{CF-Route-Datasets} by only 1-2\%.
Rather, MF1 increases slightly in some VRPs.
This generalization ability for CF routes demonstrates that our model can be applied to both actual and CF routes in our framework.\vspace{1.5ex plus .2ex minus .2ex}

\noindent\textbf{Ablation study}\ \ 
In terms of ablation studies of model architecture, we compare EC$^{\text{--enc}}_{\text{scbce}}$ and EC$^{\text{--dec}}_{\text{cbce}}$\footnote{EC$^{\text{--dec}}_{\text{cbce}}$ employs CBCE instead of SCBCE, since MLP does not consider sequence.}.
Table \ref{tab:results} shows that the tendency of the drop of the macro f1 score by the ablation varies among VRPs.
In TSPTW, EC$^{\text{--enc}}_{\text{scbce}}$ drops 0.5-1.8 \% from EC$_{\text{scbce}}$, whereas EC$^{\text{--dec}}_{\text{cbce}}$ drops 1.4-3.3 \% from EC$_{\text{cbce}}$.
In the other VRPs, EC$^{\text{--enc}}_{\text{scbce}}$ drops 5.4-11.5 \% from EC$_{\text{scbce}}$, whereas EC$^{\text{--dec}}_{\text{cbce}}$ drops 1.8-5.5 \% from EC$_{\text{cbce}}$.
These results show that the significance of the encoder and decoder increases in the other VRPs, in which the number of visited nodes per route varies (Fig. \ref{fig:statistics}).
Furthermore, the significance of the decoder outweighs that of the encoder in TSPTW, while the tendency is the opposite in the other VRPs.

\begin{figure*}[t] \centering
    \includegraphics[width=\textwidth]{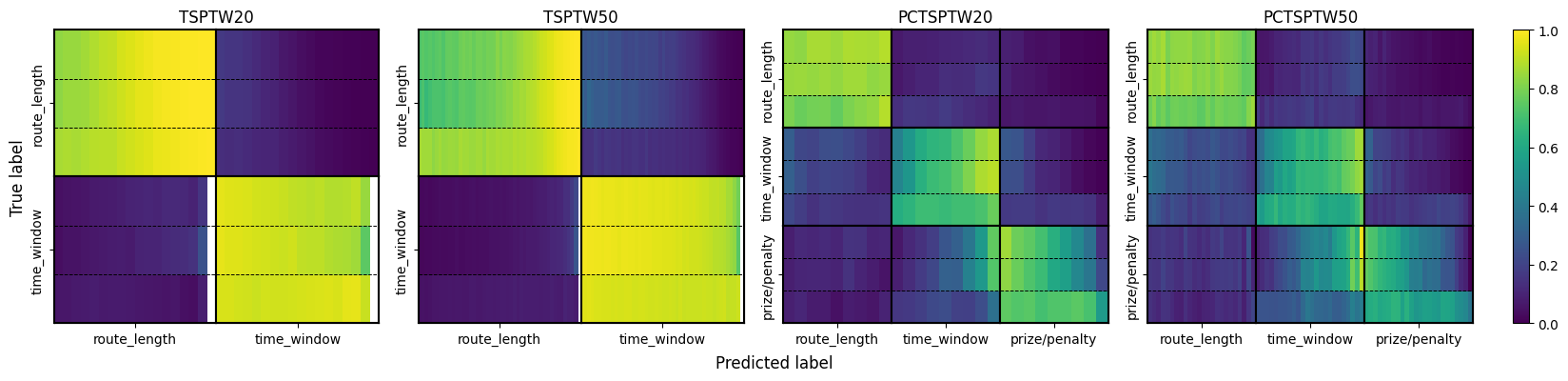}
    \caption{Sequential confusion matrices that consist of $C\times C$ grids, where each grid visualizes the normalized confusion matrix value at each step as a heatmap in the order of steps from left to right. The top, middle, and bottom in each grid correspond to EC$_{\text{ce}}$, EC$_{\text{cbce}}$, and EC$_{\text{scbce}}$, respectively.}
    \label{fig:temp_confmat}
\end{figure*}

For ablation studies of loss functions, we compare EC$_{\text{ce}}$, EC$_{\text{cbce}}$, and EC$_{\text{scbce}}$.
Table \ref{tab:results} shows that MF1 of EC$_{\text{scbce}}$ slightly drops compared to the other two models on most datasets.
However, it is insufficient to evaluate models based solely on a averaged metric (i.e., MF1), particularly in the presence of step-wise class imbalances.
Fig. \ref{fig:temp_confmat} shows the sequential confusion matrix \cite{temp_confmat} of the three models on TSPTW and PCTSPTW with $N=20, 50$. We observe that the error tendency of EC$_{\text{ce}}$ and EC$_{\text{cbce}}$ strongly depends on the step-wise class imbalance shown in Fig. \ref{fig:statistics}, i.e., they tend to fail to classify the minority class at each step. On the other hand, EC$_{\text{scbce}}$ successfully reduces the misclassification of the minority class while retaining reasonable accuracy for the other classes. 
Therefore, it is advisable to employ EC$_{\text{scbce}}$ in the presence of step-wise class imbalances, while EC$_{\text{ce}}$ is more suitable when such imbalances are absent.

\subsection{Qualitiative Evaluation of Generated Explanations}
\label{subseq:quality_eval}
\begin{figure*}[t] \centering
    \includegraphics[width=\textwidth]{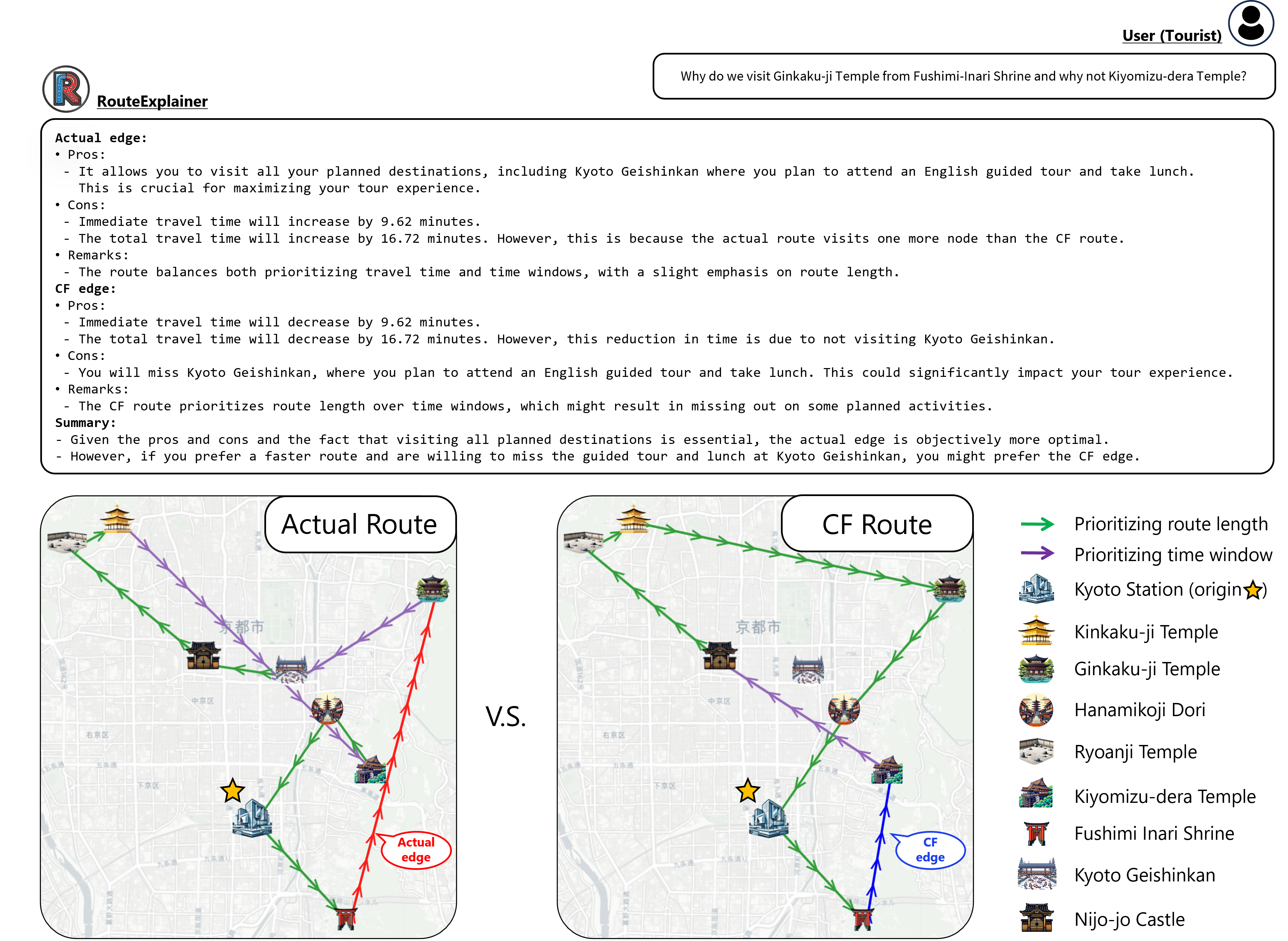}
    \caption{An explanation generated by our framework in the case study of TSPTW.} 
    \label{fig:explanation}
\end{figure*}

In this section, we qualitatively evaluate explanations generated by our framework. 
As a case study in TSPTW, we consider a tourist route that visits historical buildings in Kyoto within their time windows.
Each destination includes information about a user-defined time window, duration of stay, and remarks (e.g., take lunch).
The travel time between two destinations is calculated using Google Maps\footnote{https://developers.google.com/maps/}.
Fig. \ref{fig:explanation} shows the generated explanation text with the visualization of the actual and CF routes.
Here, a user tries to understand the influence of the actual edge from \textit{Fushimi Inari Shrine} to \textit{Ginkaku-ji Temple} so that the user considers whether the actual edge could be replaced with the user's preference (i.e., the CF edge towards \textit{Kiyomizu-dera Temple}). 
The generated explanation successfully describes the importance of the actual edge while mentioning each component in Eq. (\ref{eq:explanation}). 
The visualization of the intentions of each edge helps the user understand the immediate effect of each edge. 
Thanks to LLMs, the explanation succeeds in detailing the losses from not visiting the \textit{Kyoto Geishinkan}, incorporating the remarks.
Overall, the generated explanation assists the user in making decisions to edit the edges of the automatically generated route.
See Appendix \ref{appendix:llm} for the details of destinations.

\section{Conclusion and Future Work}
In this paper, we proposed a post-hoc explanation framework that explains the influence of each edge in a VRP route.
We introduced EIM, a novel causal model, and the pipeline of generating counterfactual explanations based on EIM.
Furthermore, we enhanced the explanation with the classifier predicting the intentions of each edge and LLM-powered explanation generation.
Through quantitative evaluation of the classifier and qualitative evaluation of generated explanations, we confirmed the validity of our framework and the effectiveness of LLMs in explanation generation.
We hope this paper sheds light on both explainability in VRP and the combination of explanation frameworks and LLMs. 

In future work, we will address two current problems:  inadequate classification performance for some VRPs and the limitation of the rule-based edge annotation for more complicated VRPs. We will address them by increasing training datasets and parameters of our model and leveraging LLM-powered annotation.

%
%
%
\bibliographystyle{splncs04}
\bibliography{ref}
\clearpage
\appendix
\section*{Appendices}
\section{Transformer Encoder Layer}
In this section, we describe the Transformer encoder layers used in the edge classifier \cite{transformer}. The one layer consists of a multi-head self-attention (MHA) layer followed by a feed-forward network (FFN). The MHA layer first computes the query $\bm{q}_{im}=W^Q_m\bm{h}_{{v_i}}$, key $\bm{k}_{im}=W^K_m\bm{h}_{{v_i}}$, and value $\bm{v}_{im}=W^V_m\bm{h}_{{v_i}}$, where $W^K_m, W^Q_m, W^V_m\in\mathbb{R}^{\frac{H}{M}\times H}$ are the $m$-th head's projection matrices, $M$ is the number of heads, and $\bm{h}_{v_i}$ is the token of $i$-th node.
It then computes the scaled dot-product attention $a_{ijm}$ between $v_i$ and $v_j$ using their key and query and aggregates the values of $v_j$ weighted by the attention coefficients. 
\begin{equation}
    \begin{aligned}
        \label{eq:attention}
        \bm{\tilde{h}}_{{v_i}m} = \sum_{j=1}^N a_{ijm}\bm{v}_{jm},\ 
        a_{ijm} =\textrm{Softmax}\left(\frac{\bm{q}_{im}^\top\bm{k}_{jm}}{\sqrt{d_k}}\right).
    \end{aligned} 
\end{equation}
After the aggregation, it concatenates the $M$ heads and computes linear projection. Hence the MHA layer is formulated as,
\begin{equation}
    \textrm{MHA}_{v_i}\left(\bm{h}_{v_1},\ldots,\bm{h}_{v_N}\right) = \sum_m W_m^O\bm{\tilde{h}}_{{v_i}m}, 
\end{equation}
where $W^O_m\in\mathbb{R}^{H\times\frac{H}{M}}$ is the projection matrix. The subsequent FFN consists of two linear transformations with a ReLU activation in between. 
With a residual connection \cite{residual} and layer normalization (LN) \cite{ln} in the outputs of MHA layers and FFNs, we obtain the embeddings $\bm{h}^{(l)}_{v_i}$ in the $l$-th layer.  
\begin{align}
    \bm{\hat{h}}^{(l)}_{v_i} &= \textrm{LN}\left(\bm{h}^{(l-1)}_{v_i} + \textrm{MHA}_{v_i}\left(\bm{h}^{(l-1)}_{v_1},\ldots,\bm{h}^{(l-1)}_{v_N}\right)\right), \\
    \bm{h}^{(l)}_{v_i} &= \textrm{LN}\left(\bm{\hat{h}}^{(l)}_{v_i} + \textrm{FFN}(\bm{\hat{h}}^{(l)}_{v_i})\right) \eqqcolon \textsc{Xfmr}_{v_i}\left(\bm{h}_{v_1},\dots,\bm{h}_{v_N}\right).
\end{align}
You may apply this Trasformer encoder layer to the sequence of edges by replacing the token of $i$-th node $\bm{h}_{v_i}$ with that of $t$-th edge $\bm{h}_{e_t}$.

\section{Data generation}
\label{appendix:data_generation}
In this section, we describe the rule-based edge annotation and the details of datasets.
\subsection{\textbf{The rule-based edge annotation}}
Algorithm \ref{alg:annotation} describes the edge annotation, which takes a route whose edges will be annotated, the VRP instance where the input route is solved, a VRP solver, and compared problems. The compared problems are the simplified VRPs that have some objectives or constraints removed from the VRP where the input route is solved.
To annotate the edge $e_t$, it first solves the instance in the compared problems while fixing the edges that are before step $t$ in the input route (line 6).
Then it compares the edge $e_t$ with the edges at step $t$ in the routes of the compared problems (line 7).
If a pair of identical ones exists, the edge $e_t$ is labeled with the corresponding problem number $c$, otherwise $C-1$ (line 8, 11).
For TSPTW, we compare the TSPTW route with the TSP route, i.e., $C=1$ and $\textsc{Vrp}_0=\textrm{TSP}$.
Since TSP only aims to minimize route length, if the edges at step $t$ in the TSPTW and TSP routes are identical, that edge of the TSPTW route is labeled 0 (route length priority).
Otherwise, we interpret that time constraint changes the edge of the TSP route, which is the optimal edge to minimize route length, to the edge of the TSPTW route. Therefore, we label that edge of the TSPTW route 1 (time constraint priority).

\begin{figure}[!t]
\IncMargin{1em}
\begin{algorithm}[H]
    \caption{Edge annotation algorithm}
    \label{alg:annotation}
    \SetKwInOut{Input}{Input}
    \SetKwInOut{Output}{Output}
    \Input{A route whose edges will be annotated $\bm{e}=(e_1,\ldots,e_{T-1})$; \newline 
          A VRP instance $\mathcal{G}$; \newline
          A VRP solver \textsc{Solver}; \newline 
          Compared VRPs \textsc{Vrp}$_{c=0, \ldots, C-2}$}
    \Output{Labels of the edges in the input route\newline $\bm{y}=(y_1,\ldots,y_{T-1})\in\{0,\ldots,C-1\}^{T-1}$}
    \SetNlSty{textbf}{}{:}
    $\bm{y}\leftarrow\phi;\ \bm{e}_{\textrm{fixed}}\leftarrow\phi$\\
    \For{$t=1,\ldots,T-1$} {
        \If{$t>1$} {
            $\bm{e}_{\textrm{fixed}}\leftarrow (e_1,\ldots,e_{t-1})$
        }
        \For{$c=0,\ldots,C-2$} {
            $\bm{e}'\leftarrow \textsc{Solver}(\textsc{Vrp}_c, \mathcal{G}, \bm{e}_{\textrm{fixed}})$\\
            \If{$e_t=e'_t$} {
                $y_t\leftarrow c$ \\
                \textbf{break}
            }
        }
        \If{$y_t=\phi$}{
            $y_t\leftarrow C-1$
        }
    }
    \Return $\bm{y}$
\end{algorithm}
\DecMargin{1em}
\end{figure}

\subsection{\textbf{Problems and Datasets}}
\begin{figure*}[tb] \centering
    \includegraphics[width=\textwidth]{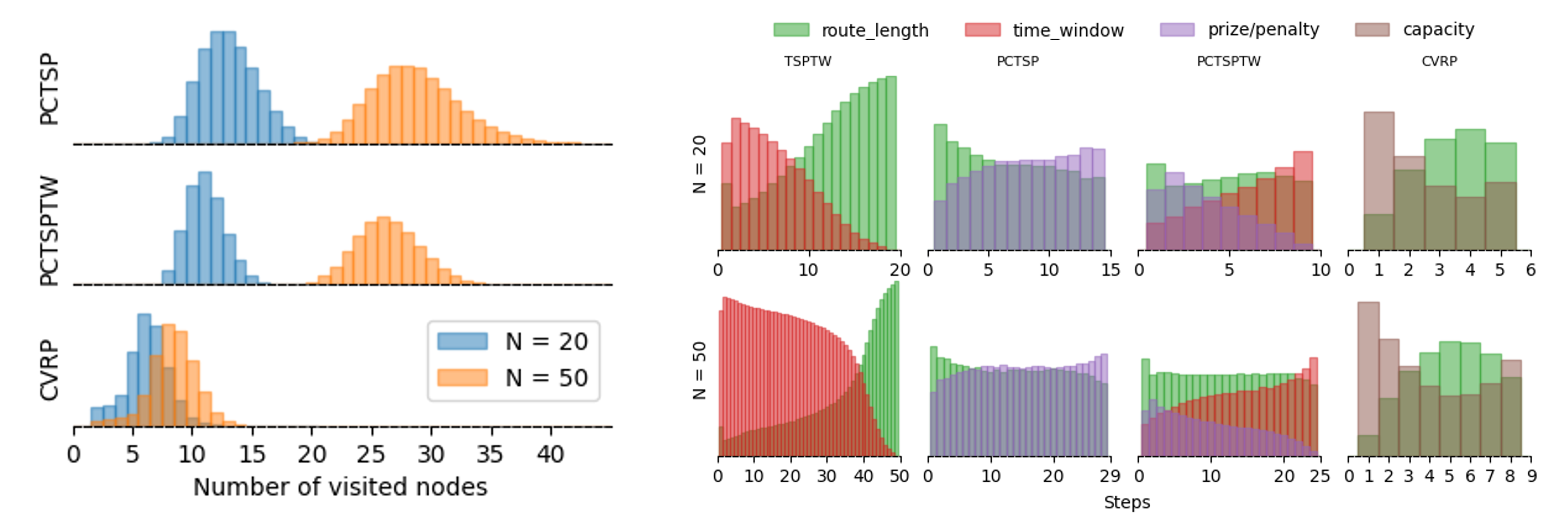}
    \caption{Distribution of the number of visited nodes per route (left) and the ratio of the classes of edges w.r.t. steps (right) on each training VRP dataset. 
    Each edge is annotated by Algorithm \ref{alg:annotation}.
    The class ratios shown in PCTSP, PCTSPTW, and CVRP are for samples in which the number of visited nodes is the mode. TSPTW and PCTSPTW show the step-wise class imbalance.} 
    \label{fig:statistics2}
\end{figure*}

We generate four different VRP datasets of which the problem size $N=20, 50$. 
Fig. \ref{fig:statistics2} and Table \ref{tab:statics} show the statistics of each training dataset.
In the following, we describe the generation process of the actual route datasets in each VRP.
\begin{table}[tb]
    \caption{The number of edges and overall class ratio on the training datasets (128K instances). The class ratio is shown from left to right following the order of route\_length, time\_window, prize/penalty, capacity.}
    \begin{center}
    \begin{tabular}{l @{\extracolsep{4pt}}ccll@{}}
        \hline
        &\multicolumn{2}{c}{\# edges} &\multicolumn{2}{c}{class ratio (\%)} \\
        \cline{2-3}\cline{4-5}
        &$N=20$ &$N=50$ &\ \ \ \ \ \ $N=20$ &\ \ \ \ \ \ $N=50$ \\
        \hline
        TSPTW   &24M  &63M  &61.9 : 38.1 &35.0 : 65.0 \\
        PCTSP   &15M  &35M  &50.9 : 49.1 &50.2 : 49.8 \\
        PCTSPTW &13M  &33M  &43.8 : 33.6 : 22.6  &47.5 : 35.7 : 16.8 \\
        CVRP    &26M  &64M  &56.1 : 43.9  &48.3 : 51.7 \\
        \hline
    \end{tabular}
    \label{tab:statics}
    \end{center}
\end{table}\vspace{1.5ex plus .2ex minus .2ex}

\noindent\textbf{TSP with Time Window (TSPTW)}\ \ 
To ensure the existence of the classes of both route length priority and time window priority, we generate the instances based on the \textit{large time window dataset} \cite{dpdp}, where the node features are sampled as follows: the coordinates $\bm{x}\sim\mathcal{U}(0, 1)$, the earliest time $e_i\sim\mathcal{U}(t_i^\text{arr}-\frac{t_{\text{max}}}{2}, t_{i}^\text{arr})$, the latest time $l_i\sim\mathcal{U}(t_i^\text{arr}, t_i^\text{arr} + \frac{t_{\text{max}}}{2})$. 
$\mathcal{U}$ denotes uniform sampling, $t_i^\text{arr}$ is the time at which the $i$-th node is visited on a randomly generated route, and $t_\text{max}$ is the maximum time window (we set $t_\text{max}=10$).
We solve the instances with OR-Tools and annotate each edge in the generated route with Algorithm \ref{alg:annotation} using LKH as the VRP solver.
Note that we skip instances that cannot be solved within a specific time limit and repeat the above process until the desired number of instances is reached. The input features for the edge classifier are the coordinates and time windows. The global state is the currently accumulated travel time.\vspace{1.5ex plus .2ex minus .2ex}

\noindent\textbf{Prize Collecting TSP (PCTSP)}\ \ 
In the PCTSP, each node has both its prize and penalty. The goal here is to find a route that collects at least a minimum total prize while minimizing both the route length and the total penalty of unvisited nodes. We use the datasets by \cite{kool2018attention} as the instance generation, where node features are sampled as follows: the coordinates $\bm{x}\sim\mathcal{U}(0, 1)$, the prize $\rho_i\sim\mathcal{U}(0, \frac{4}{N})$, the penalty $\beta_i\sim\mathcal{U}(0, 3\cdot\frac{K}{N})$, where $N$ is the number of nodes and $K=2, 3$ for $N=20, 50$.
The state values in Eq. (\ref{eq:edge_encoder}) are here the total prize and penalty at $t$.
We solve the instances with OR-Tools and annotate each edge in the generated route with LKH.
In the edge annotation, we compare the PCTSP route with the TSP one, where an edge is labeled route length priority if the edges in both routes match, otherwise, prize/penalty priority. The input features for the edge classifier are the coordinates, prizes, and penalties. The global states are the currently accumulated prize and penalty.\vspace{1.5ex plus .2ex minus .2ex}

\noindent\textbf{Prize Collecting TSPTW (PCTSPTW)}\ \ 
The PCTSPTW adds a time constraint to the PCTSP that each node must be visited within its time window.
We generate the instances by combining the procedures of the PCTSP and TSPTW.
One change here is that we increase the time window size as $t_\text{max}=50$ for $N=50$.
We solve the instances and annotate each edge in the generated route with OR-Tools (because LKH does not support PCTSPTW).
In the edge annotation, we compare the PCTSPTW, TSPTW, and TSP routes, where the edge is labeled route length priority if the edges in the PCTSPTW and TSP routes match, time constraints priority if the edges in the PCTSPTW and TSPTW routes match, otherwise, prize/penalty priority.
The input features for the edge classifier are the coordinates, prizes, penalties, and time windows. The global states are the currently accumulated prize, penalty, and travel time.\vspace{1.5ex plus .2ex minus .2ex}

\noindent\textbf{Capacitated VRP (CVRP)}\ \ 
In the CVRP, each node has a demand, and the vehicle capacity is given.
The goal is to find a set of routes that minimizes the total route length, where each route should start and end at the depot, and the total demands of nodes visited in each route do not exceed the vehicle capacity. We generate the instances based on the dataset by \cite{nazari}, where
the coordinates $\bm{x}\sim\mathcal{U}(0, 1)$, the demands $\delta_i\sim\mathcal{U}([1,\dots,9])$, and the vehicle capacity $cap=20, 40$ for $N=20, 50$.
The state value in Eq. (\ref{eq:edge_encoder}) is the remaining capacity at $t$.
We solve the instances with OR-Tools and annotate each edge in the generated route with LKH.
In the edge annotation, we compare each CVRP route with the TSP one, where the edge is labeled route length priority if the edges in both routes match, otherwise, capacity priority. 
The input features for the edge classifier are the coordinates and demand. The global state is the currently remaining capacity.\vspace{1.5ex plus .2ex minus .2ex}

\noindent\textbf{CF Route Datasets}\ \ 
Each CF route dataset includes 10K sets of CF routes and labels of their edges, where the CF routes are generated with the CF edge uniformly sampled in the routes of the actual route datasets for evaluation. In each VRP, we use the same VRP solver as in the actual route dataset for generating the CF routes and annotating their edges.

\section{Details of Baselines}
\label{appendix:baslines}
To qualify why we need the edge classifier instead of Algorithm \ref{alg:annotation}, we compare the edge classifier with Algorithm \ref{alg:annotation} that uses different VRP solvers. The quality of generated routes and the calculation time differ among each VRP solver. One that provides the most near-optimal routes among these solvers is the annotator (i.e., ground truth), and others are baselines.\vspace{1.5ex plus .2ex minus .2ex}

\noindent\textbf{Google OR-Tools (OR-Tools)}\ \ 
OR-Tools is an open-source solver that supports various optimization problems, including VRP.
We set its parameters as follows: the search time limit is 5 seconds; the first solution strategy is Path Cheapest Arc; the metaheuristic is Guided Local Search; the other parameters are default.
As OR-Tools takes only integer inputs, we use node features that are multiplied by a large value (e.g., 1e+6) and rounded to integers, which is the same as in LKH and Concorde. To force it to include specified edges in the solution, we use a built-in function \texttt{ApplyLocksToAllVehicles}.\vspace{1.5ex plus .2ex minus .2ex}

\noindent{\textbf{Lin-Kernighan-Helsgaun (LKH)}\ \ 
LKH \cite{lkh} is a VRP solver that uses variable depth local search of LK.
We set its parameters as follows: the total number of runs is 1; the maximum number of trials per run is 10; \texttt{PATCHING\_A} is 2; \texttt{PATCHING\_C} is 3; the other parameters are default. To force it to include specified edges in the solution, we set \texttt{FIXED\_EDGE\_SECTION} in the TSPLIB file.\vspace{1.5ex plus .2ex minus .2ex}

\noindent{\textbf{Concorde TSP Solver (Concorde)}\ \ 
Concorde is a solver specialized for TSP. We keep all parameters default.
As Concorde does not support generating a route that includes specified edges, we realize that with an alternative way: we set weights of the specified edges to zero and set weights of edges between visited and unvisited nodes to a large value. However, this implementation would have a limitation/bug in that the solver outputs no route when the number of unvisited nodes is small. This behavior was observed in CVRP with $N$=20 (Table \ref{tab:results}).

\section{Hyperparameters and Devices}
\label{appendix:hyperparameter}
We initialize trainable parameters with $\mathcal{U}(-1/\sqrt{d}, 1/\sqrt{d})$, $d$ is the input dimension. 
For each VRP and problem size, we prepare 128K, 10K, and 10K instances for training, validation, and evaluation, respectively. 
The maximum epoch is 100, and (mini-batch size, constant learning rate) is (512, $10^{-4}$) for CVRP with $N=50$ and (256, $10^{-3}$) for other VRPs. We set the parameter of the class-balanced loss $\beta=0.99$. For the Transformer encoder layer in the node encoder and decoder, we set the number of layers $L=L'=2$, the dimension in hidden layers $H=128$, and the number of heads $M=8$.
We set the temperature of GPT-4 to 0 for improving consistency of responses.
We commonly use the random seed 1234 in the entire process.
We conducted the experiments once on a single GPU (NVIDIA A100: 80G) and two CPUs (AMD EPYC 7453: 28 cores, 56 threads).

\section{Implementation Details}
\label{appendix:llm}
\textbf{Overall Algorithm}\ \ 
Here, we wrap up our framework by describing its overall flow (Algorithm \ref{alg:framework}).
The framework takes a why and why-not question, the VRP instance associated with the actual route, a VRP solver, a trained edge classifier, functions that compute and compare representative values, and an LLM.
First, it derives the CF route for the why and why-not question (lines 1, 2).
Second, for the actual and CF routes, it classifies the edges in both routes with the trained edge classifier and then computes the representative values of the actual and CF edges (line 3-6).
Lastly, the LLM generates a counterfactual explanation by comparing the representative values of the actual and CF edges (line 7, 8).
\begin{figure}[!t]
\IncMargin{1em}
\begin{algorithm}[H]
    \caption{RouteExplainer}
    \label{alg:framework}
    \SetKwInOut{Input}{Input}
    \SetKwInOut{Output}{Output}
    \Input{A why and why-not question $\mathcal{Q}=(\bm{e}^\text{fact}, t_\text{ex}, e^\text{fact}_{t_\text{ex}}, e^\textrm{cf}_{t_\text{ex}})$; \newline
           A VRP instance $\mathcal{G}$; \newline
           A VRP solver $\textsc{Solver}$; a trained edge classifier $f_\text{classifier}$;\newline
           A set of representative value functions $\mathcal{F}_\text{rep}$;\newline
           A set of compare functions $\mathcal{F}_\text{compare}$;\newline
           The VRP where the actual route is solved \textsc{Vrp}; \newline
           An LLM \textsc{Llm}}
    \Output{A counterfactual explanation written in natural language}
    \SetNlSty{textbf}{}{:}
    $\bm{e}_{\textrm{fixed}}\leftarrow (e^{\text{fact}}_1,\ldots,e^{\text{fact}}_{t_\text{ex}-1}, e^\textrm{cf}_{t_\text{ex}})$\\
    $\bm{e}^{\mathcal{Q}}\leftarrow \textsc{Solver}(\textsc{Vrp}, \mathcal{G}, \bm{e}_{\textrm{fixed}})$\\
    \For{$k\in\{ \rm{fact}, \mathcal{Q} \}$} {
        $\bm{\hat{y}}^k\leftarrow f_\text{classifier}(\bm{e}^k, \mathcal{G})$\\
        $\bm{I}_{e^k_{t_\text{ex}}}\leftarrow\left(S_{(\pi^k_{t+1}, t+1)}, \hat{y}^k_{t_\text{ex}+1}, S_{(\pi^k_{t+2}, t+2)}, \ldots, \hat{y}^k_{T-1}, S_{(\pi^k_{T}, T)}\right)$ \\
        $\mathcal{R}_{e^k_{t_\text{ex}}}\leftarrow \mathcal{F}_\text{rep}(\bm{I}_{e^k_{t_\text{ex}}})$
    }
    $\mathcal{X}_\mathcal{Q}\leftarrow \mathcal{F}_\text{compare}(\mathcal{R}_{e^\text{fact}_{t_\text{ex}}}, \mathcal{R}_{e^\mathcal{Q}_{t_\text{ex}}})$ \\
    \Return \textsc{Llm}($\textsc{Vrp}, \mathcal{X}_\mathcal{Q}$)\\
\end{algorithm}
\DecMargin{1em}
\end{figure}\vspace{1.5ex plus .2ex minus .2ex}

\noindent\textbf{System Architecture}\ \ 
Fig. \ref{fig:system_arc} shows the system architecture for our framework, which has two separate processes: route generation and explanation generation. 
In route generation, the system takes a set of destinations given by a user and provides a near-optimal route that visits the destinations (i.e., actual route). In explanation generation, receiving the actual route, the user first asks a why and why-not question for an edge in the route in natural language text. The question text and the information on the actual route are then embedded in a system prompt (Fig. \ref{fig:input_extractor_prompt}). An LLM takes the prompt and identifies which edges correspond to the actual and CF edges in the user's question. LLM then generates the CF route by calling the CF generator with the corresponding arguments. After giving the intentions to edges in the actual and CF routes by the edge classifier, information so far is embedded in another system prompt (Fig. \ref{fig:explanation_prompt}). Lastly, the LLM takes this prompt and generates explanation text.
We employ GPT-4 as the LLM in our framework and implement the pipeline with LangChain\footnote{\url{https://github.com/langchain-ai/langchain}}.

\begin{figure*}[tb] \centering
    \includegraphics[width=\textwidth]{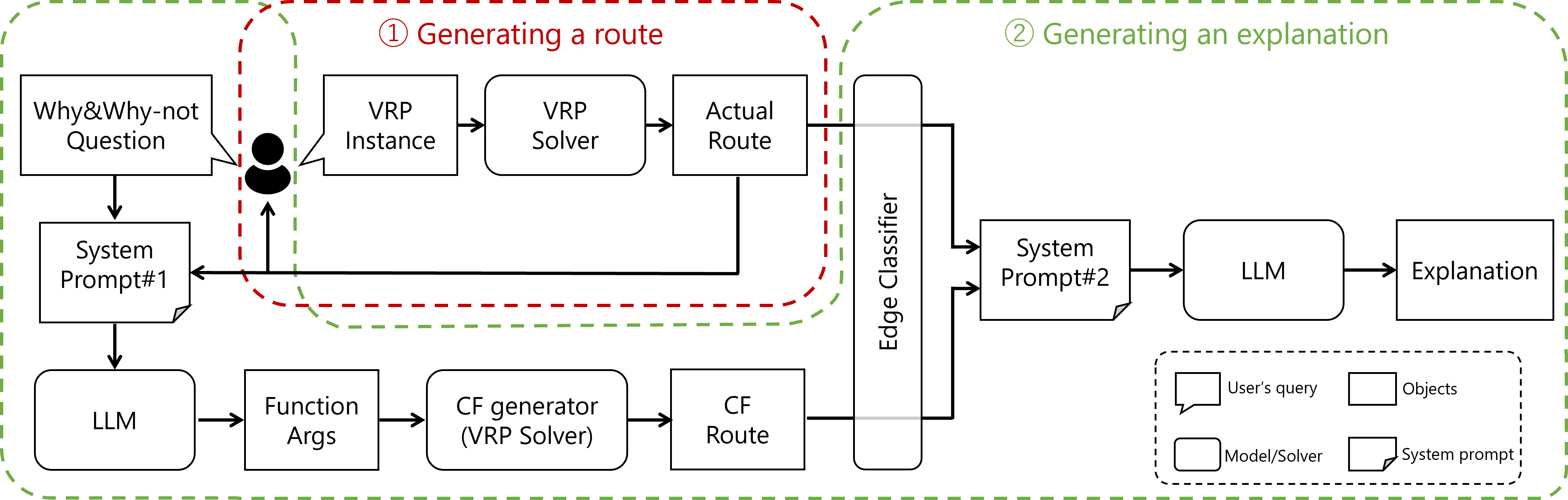}
    \caption{System architecture implementing our framework} 
    \label{fig:system_arc}
\end{figure*}
\begin{figure*}[tb] \centering
    \includegraphics[width=\textwidth]{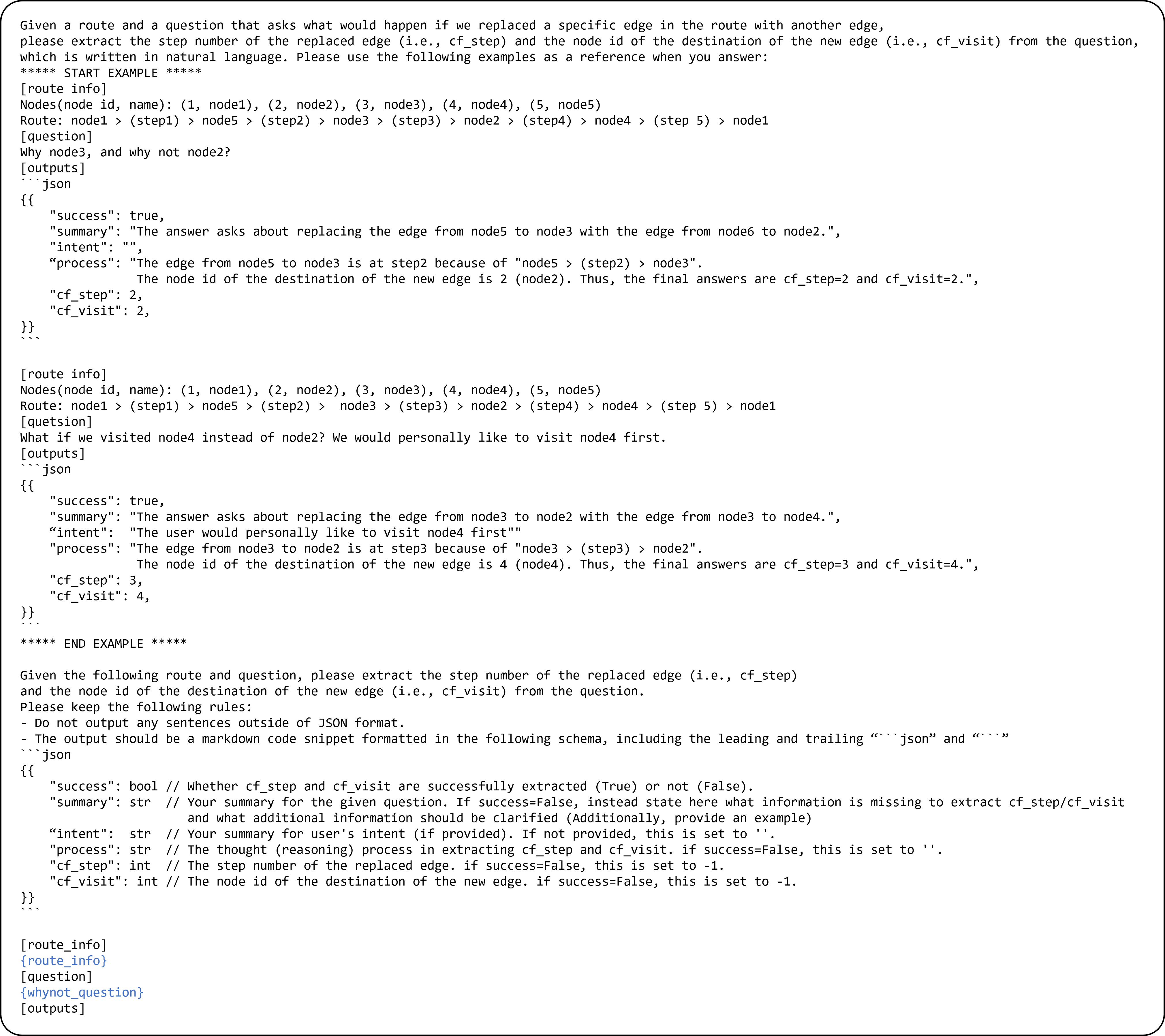}
    \caption{System prompt\#1: a system prompt for extracting input arguments from the user's question written in natural language. The information of the generated route, formatted as in the example, and the user's question text are respectively embedded in \texttt{\{route\_info\}} and \texttt{\{whynot\_question\}}. This template enables LLMs to identify $t_\textrm{ex}$ and $e_{t_\textrm{ex}}^\textrm{cf}$ within the question text by referring to the route information. The output is specified in JSON format, and by parsing the JSON output, we can obtain the \texttt{dict} of the arguments needed for the CF generator.} 
    \label{fig:input_extractor_prompt}
\end{figure*}
\begin{figure*}[tb] \centering
    \includegraphics[width=\textwidth]{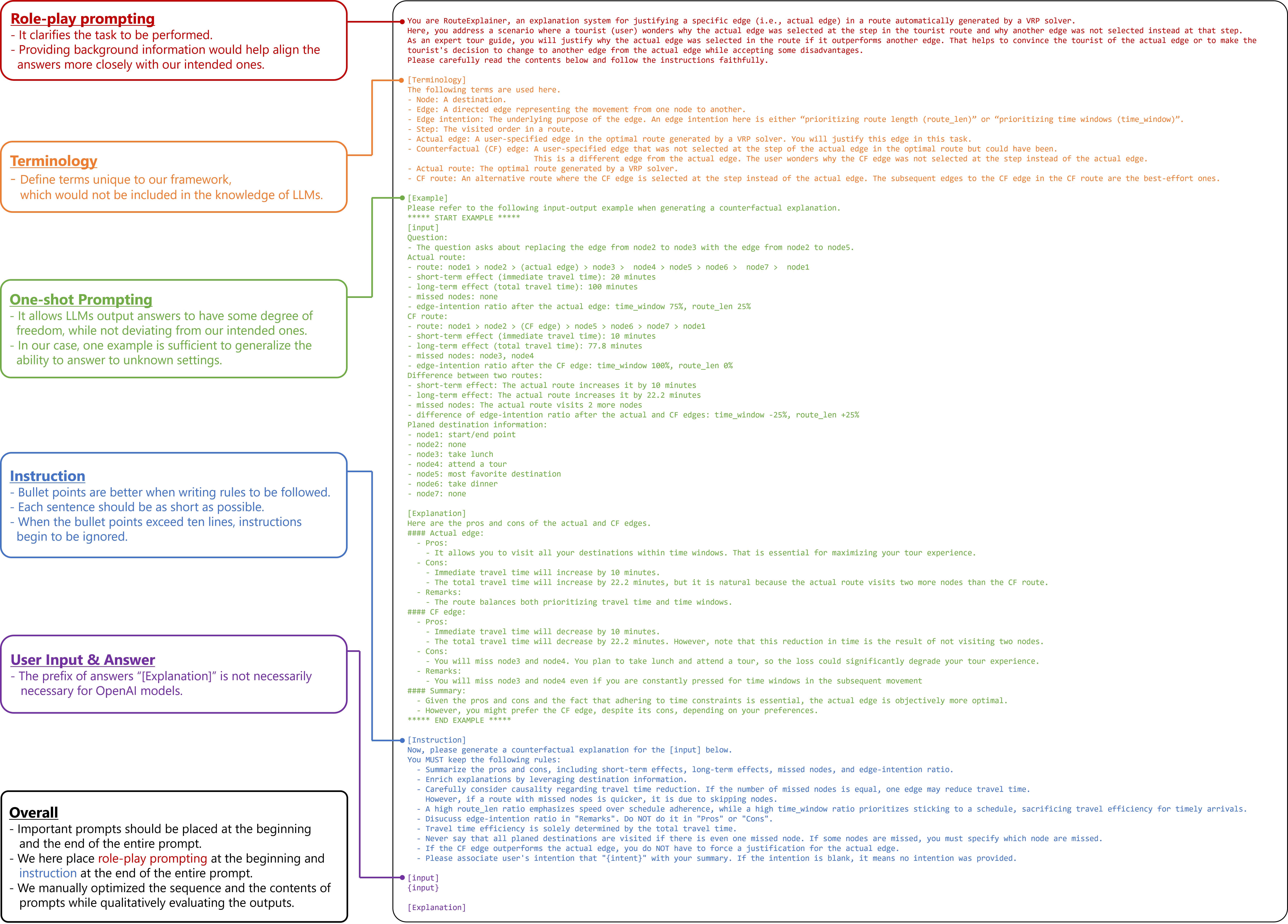}
    \caption{System prompt\#2: a system prompt for generating a counterfactual explanation based on our framework. The comparison result between the actual and CF edges, formatted as in the example, is embedded in \texttt{\{input\}}. This prompt comprises four components: specifying a role, describing terminology, presenting an example Q\&A, and instructions. Role-play prompting is a well-known technique to improve the capability of LLMs in various settings \cite{roleplay1,roleplay3,roleplay2} . One-shot prompting of Q\&A is employed to allow LLMs output answers to have some degree of freedom while not deviating from our intended ones. Overall, we optimized the order of prompts so that the more important ones come first and last \cite{llmlost}. Note that we manually optimized this system prompt by qualitatively evaluating the responses, rather than through systematic optimization.} 
    \label{fig:explanation_prompt}
\end{figure*}\vspace{1.5ex plus .2ex minus .2ex}

\noindent\textbf{Application Demo}\ \ 
Fig. \ref{fig:app} shows an application demo for our framework, which is made with Streamlit\footnote{\url{https://github.com/streamlit/streamlit}}. 
A user first determines destinations, their time windows, and remarks. In this demo, all destinations are pre-determined, of which contents are shown in Table \ref{tab:tour_list}. After generating a TSPTW route that visits the input destinations, the user asks a why-not question in the chat box, e.g., \textit{Why do we visit Ginkaku-ji Temple after Fushimi-Inari Shrine, instead of Kiyomizu-dera?}
The app then displays the explanation text with the visualization of the actual and CF routes.
Lastly, the user has the option of keeping the actual route or replacing it with the CF edge, where the user selects one of them while understanding their pros and cons based on the explanation.
By iteratively repeating this process, the user can generate their preferred route. 
Note that this route is optimal for the user's preferences but may not be mathematically optimal.

The benefit of using natural language inputs lies in the user's ability to specify both the actual and CF edges intuitively. Additionally, this allows for the future expansion of the interface to include voice input.

\begin{table}[tb]
    \caption{Destination list commonly used in the qualitative evaluation and the application demo. Open-close indicates a time window, and we must \textit{arrive at each node within its time window}, i.e., (arrival\_time + stay\_duration) $>$ close\_time is allowed.}
    \scriptsize
    \begin{center}
    \begin{tabular}{lcccl}
        \hline
        destination &open\_time &close\_time &stay\_duration (h) &remarks \\
        \hline
        Kyoto Station         &7:00  &22:00  &0 &Start/end point \\
        Kinkaku-ji Temple     &9:00  &17:00  &1 &None \\
        Ginkaku-ji Temple     &9:00  &16:30  &1 &November schedule \\
        Fushimi-Inari Shrine  &8:30  &16:30  &1 &For prayer \\
        Kiyomizu-dera Temple  &6:00  &18:00  &1 &None \\
        Nijo-jo Castle        &8:45  &16:00  &1 &None \\
        Kyoto Geishinkan      &10:30 &11:30  &2.5 &Attend an English guided tour and take lunch \\
        Ryoanji Temple        &8:30  &16:30  &1 &None \\
        Hanamikoji Dori       &19:00 &20:00  &1 &Take dinner \\
        \hline
    \end{tabular}
    \label{tab:tour_list}
    \end{center}
\end{table}

\begin{figure*}[tb] \centering
    \includegraphics[width=0.9\textwidth]{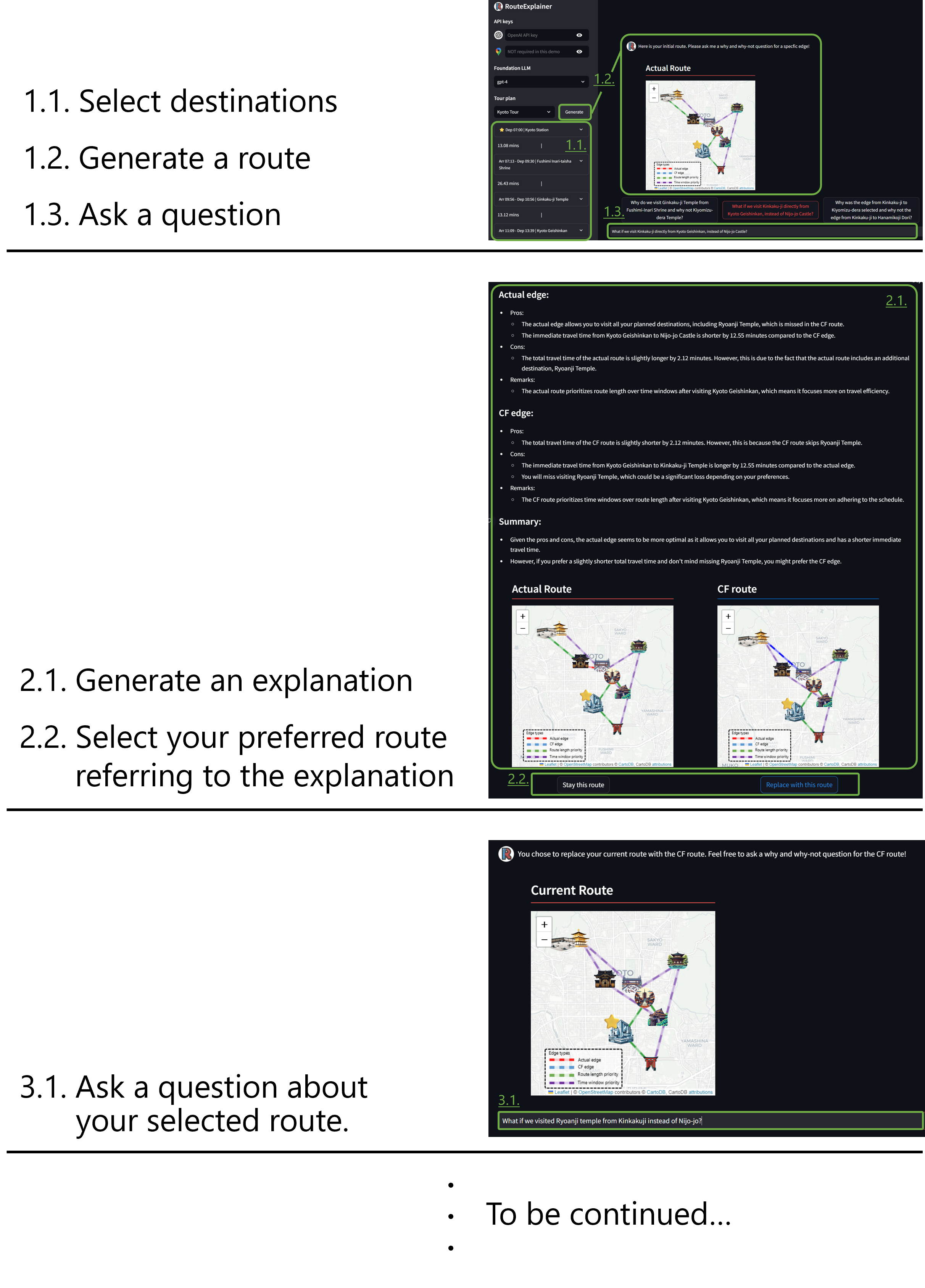}
    \caption{Screenshots of the application demo for our framework.} 
    \label{fig:app}
\end{figure*}


\end{document}